\documentclass{article}

\usepackage[final]{neurips_2020}

\usepackage[utf8]{inputenc} 
\usepackage[T1]{fontenc}    
\usepackage{hyperref}       
\usepackage{url}            
\usepackage{booktabs}       
\usepackage{amsfonts}       
\usepackage{nicefrac}       
\usepackage{microtype}      
\usepackage{caption,graphicx}
\usepackage{subcaption}
\usepackage{xcolor}
\usepackage{graphicx,wrapfig,lipsum}
\usepackage{multirow}

\definecolor{darkgreen}{HTML}{036400}  
\definecolor{darkred}{HTML}{680100} 
\definecolor{OliveGreen}{cmyk}{0.64,0,0.95,0.40}

\title{The Lab vs The Crowd: An Investigation into Data Quality for Neural Dialogue Models}

\author{%
  Jos{\'e} Lopes, 
  Francisco J. Chiyah Garcia, 
  Helen Hastie \\
  School of Mathematical and Computer Sciences \\
  Heriot-Watt University \\
  Edinburgh, UK \\
  \texttt{\{jd.lopes, fjc3, h.hastie\}@hw.ac.uk} \\
}
  
\begin{document}

\maketitle

\begin{abstract}
 Challenges around collecting and processing quality data have hampered progress in data-driven dialogue models. 
 Previous approaches are moving away from costly, resource-intensive lab settings, where collection is slow but where the data is deemed of high quality. The advent of crowd-sourcing platforms, such as Amazon Mechanical Turk, has provided researchers with an alternative cost-effective and rapid way to collect data. 
 However, the collection of fluid, natural spoken or textual interaction can be challenging, particularly between two crowd-sourced workers. In this study, we compare the performance of dialogue models for the same interaction task but collected in two different settings: in the lab vs. crowd-sourced. We find that fewer lab dialogues are needed to reach similar accuracy, less than half the amount of lab data as crowd-sourced data.. We discuss the advantages and disadvantages of each data collection method. 
\end{abstract}


\section{Introduction}

 Data collection is an essential part of the field of spoken dialogue systems and conversational AI. 
 In particular, designing a dialogue system for a completely new domain is still a very challenging task.  Data collection options include running lab-based experiments, crowd-sourced tasks (e.g.  Amazon Mechanical Turk) or gathering data from social media platforms, such as Reddit or Twitter. Ambitious large scale data collections across multiple domains have resulted in widely used datasets, such as MultiWOZ \citep{budzianowski2018multiwoz}. 
 However, starting off in a new domain from scratch still has its challenges. Difficult and costly decisions have to be made as to how and where to collect the data.
 
 A large majority of recent dialogue corpora has been collected using crowd-sourcing either by pairing workers and letting them chat, often about a given topic \citep{he-etal-2017-learning}, or by asking them to add the next utterance to the dialogue given a set of conditions \citep{wen-etal-2017-network}. Other studies have recruited subjects to play the role of the system, i.e., to act as a wizard or user \citep{ElAsri2017}. Each of these approaches has its own advantages and disadvantages, depending on if the dialogue is task-oriented or not. By letting users type in an unrestricted way, the richness of the dialogue increases, which is a positive feature for chit-chat. On the other hand, too much variability could be a problem for a high stakes, task-oriented dialogues, such as in the medical domain. Letting multiple users contribute with one utterance per dialogue \citep{wen-etal-2017-network}, speeds up the data collection, however, dialogues may lack coherence and severely diverge from real dialogues. On the other hand, hiring and training subjects to chat or perform the wizard role results in a more controlled data collection but dramatically increases the cost of the data collection and makes it less scalable. 
 
 The quality of such datasets has been often assessed according to the degree of variability  observed \citep{Byrne2019} or the lexical complexity of the utterances collected \citep{jonell2019crowdsourcing,novikova-etal-2017-e2e}. 
Furthermore, most of the above-mentioned datasets focus on increasing the size of the dataset available for dialogue research, rather than investigating the impact of the data collection strategies on the performance of the models trained. The work presented in this paper aims at highlighting the pros and cons, using a methodology to quickly leverage a robust dialogue system, minimising the cost and effort involved in the data collection process. Analyses comparing different strategies for the data collection process across various platforms have been done in the past \citep{Jonell2020Trust}, but we are not aware of a similar study for dialogue data.

The data used in this study was collected in the scope of an emergency response system to be used on an off-shore energy platform as part of the EPSRC ORCA Hub programme \citep{ORCA}. One of the collections was done using crowd-sourcing \citep{chiyah-garcia-etal-2020-crwiz}
and the second one was done in a lab using a Wizard-of-Oz setting, where participants were interacting either with a social robot or a smart speaker.
Both datasets were used to train a dialogue model using an implementation of a Hybrid Code Network \citep{williams-etal-2017-hybrid} and here we compare the results achieved by models trained on data collected by either method. To validate the use of crowd-sourced data to bootstrap a dialogue system for situated interaction, we ran experiments where we train the model on the crowd-sourced data and test it on the lab data, in order to verify if it 
achieves comparable performances with the models trained only with the lab data.

The contributions of this paper are as follows: 1) a comparison of models trained with two datasets collected in different ways but on the same task, 2) evidence that suggests that specialised dialogue tasks, such as our emergency response task, are not well covered by current pre-trained dialogue models, and 3) a set of recommendations regarding the data collection for dialogue research.\footnote{Please find code and data in: \href{https://github.com/zedavid/TheLabVsTheCrowd}{\texttt{https://github.com/zedavid/TheLabVsTheCrowd}}.}

The paper is organised as follows. Section \ref{sec:rel_work} will cover previous work related to this problem. Our experimental set-up will be introduced in Section \ref{sec:exp_set_up}, followed by the results in Section \ref{sec:results}. The paper concludes with the discussion in Section \ref{sec:dicussion} and future work and conclusions in Section \ref{sec:conclusion}.

\section{Related Work}
\label{sec:rel_work}

In this section, we first discuss a number of previously used methods for dialogue data collection and then describe studies where researchers have compared different approaches in data collection and their impact the model performance. 

\subsection{Methods for data collection}
Due to the high data demands of training dialogue models, running laboratory data collections at scale for this purpose has became impractical. One of the earliest large datasets for dialogue research was collected with the Let's Go system \citep{raux2005let}. This dataset includes 171,128 interactions of real-customers with a telephone-based agent who provided bus schedule information in Pittsburgh. A similar paradigm was used in \citet{williams2016dialog}, who collected 15K dialogues, however instead of real bus passengers, crowd-workers were paid to interact with systems over the phone. Nowadays, many data collections involve crowd-workers acting either as the wizard, as in the Wizard-of-Oz paradigm \citep{Kelley1984wizard}, or the user. However, the way they contribute might be different depending on the design of the data collection. In \citet{budzianowski2018multiwoz,wen-etal-2017-network,lee2019multi-domain,eric-etal-2017-key}, crowd-workers are shown the history of the dialogue and they are asked to add a coherent response. They are typically given some constraints, which they should take into account when acting as the wizard or agent. In addition, workers may also have to perform annotation on the previous user input, according to a set of instructions given. Although this method was mostly used for task-oriented data collections, it could also be used to collect a diverse chit-chat corpus \citep{jonell2019crowdsourcing}. A variation on this data collection method is described in \citet{andreas2020dialogflow}, where the framework would generate a set candidates for the wizards to select from, in a similar fashion as to what was done in \citep{chiyah-garcia-etal-2020-crwiz}, for the data used in this paper. 

While crowd-sourcing could be a good solution for creating large amounts of data, this method has an impact on the quality of the data, due to the rapid nature of the tasks. To increase the quality of the dataset in \citet{ElAsri2017}, instead of crowd-workers, a limited set of subjects were hired. Subjects contributed to the data collection both as the user and the wizard. They swapped roles during the data collection, which made them aware the agent was played by another human. This approach is not very scalable and therefore, to perform larger collections, a hybrid model where crowd-workers are paired with trained subjects to act as the wizard has been used by \citet{peskov2019multi}. 

Hiring subjects to act as wizards is a strategy followed to improve not only the quality and consistency of the dialogues, but also the quality of the annotations, necessary for downstream tasks. This paradigm was also used in \citet{Byrne2019}, who collected a multi-domain corpus. In addition, authors have also collected a similar amount of data using self-authored dialogues, where a single crowd-worker authored both the agent and the user utterances. Unlike what was found by \citet{jonell2019crowdsourcing} for chit-chat dialogues, authors found that self-authored dialogues were richer in content and variability. A different approach has been used in a dataset where a simulator, a probabilistic automaton, was designed to create dialogues with a variety of trajectories \citep{rastogi2019towards}. Once this structured representation of dialogue semantics is created, the crowd-workers' task is to paraphrase them into natural language.

While datasets such as these are widely used for research, they have a few limitations. Firstly, for research in spoken dialogue, their usage is limited since the datasets are mostly written, not spoken. The exception is \citet{Byrne2019}, where crowd-workers could listen to the agent response using TTS. However, even in this case, the timing aspect of spoken interaction is removed. Secondly, there is some overlap in the domains covered by these datasets (e.g. between restaurants and hotels). This raises questions about their possible usage for training models in totally new domains (e.g. health diagnosis, emergency response). 

To mitigate the first limitation, there have been some collections in the wild, such as the one presented by \citet{siegert-2020-alexa}, where an Amazon Alexa was part of an itinerant exhibition, where visitors could interact with it using voice. This dataset contains more than 7,000 dialogues in German.

\subsection{Comparisons of training models using data collected using various methods}

Given that these datasets cover a variety of different domains, studies have been conducted investigating how a model trained on a source domain performs when tested on a different target domain \citep{zhao-eskenazi-2018-zero}. Another line of research has tried to find the minimal number of dialogues from the target domain that should be used during training to achieve a competitive performance \citep{shalyminov-etal-2019-shot}. 
The Schema Guided Dialogue (SGD) dataset collected by  \citet{rastogi2019towards}  was used to train a model, which was then tested in  MutilWOZ 2.1. Both datasets were crowd-sourced, but the methodology followed was slightly different. Despite this fact, the model trained with the SGD dataset outperformed the state-of-the art at the time, in terms of joint goal accuracy in MultiWOZ 2.1. 

Recently, advances in Natural Language Processing have been made by training neural models with very large datasets and applying those models to a variety of downstream tasks. These pre-trained dialogue models have been released for general use \citep{henderson2019convert,wu2020tod,zhang2019dialogpt}, and are trained on a variety of sources, including Twitter and Reddit \citep{henderson2019convert}.
While these models can address common-sense questions and perform multi-turn reasoning, they might fail to capture specific aspects of the task in task-oriented dialogues.  \citet{wu2020tod} attempted to address this with their pre-trained model, which was trained with an ensemble of several task-oriented dialogue datasets publicly available. This BERT-based model was successfully tested in different downstream tasks including dialogue act prediction, entity extraction in dialogue and response generation.
In both \citet{wu2020tod} and \citet{rastogi2019towards}, models were trained and tested on data collected with different methods. However, their focus was rather on the overall performance of the models rather than in optimising the data collection.


\section{Experimental Set-up}
\label{sec:exp_set_up}

The task used in the data collections described here is an emergency response scenario with two participants collaborating in real time to resolve an emergency on an offshore facility \citep{HastieCHIWS2019}. It was designed as a Wizard-of-Oz experiment in which the wizard, playing an intelligent assistant, serves as the intermediary between a remote facility operator and several on-site autonomous robots. In the task there is a limited time until the facility had to be evacuated and both had to avoid this by all means, whilst keeping to specific safety procedures. As such, the task required careful collaboration to resolve the high-stakes scenario in time.

Actions taken by the participants affected a dynamic simulation running in parallel (e.g. by sending a robot to an area). There were a total of four robots (two ground and two aerial robots) that could be operated in any order with a mix of capabilities available: inspecting an area, activating a hose, opening a valve. Thus, there were several ways of resolving the emergency.

\begin{figure}
    \centering
    \begin{subfigure}[b]{0.45\textwidth}
        \includegraphics[width=\textwidth]{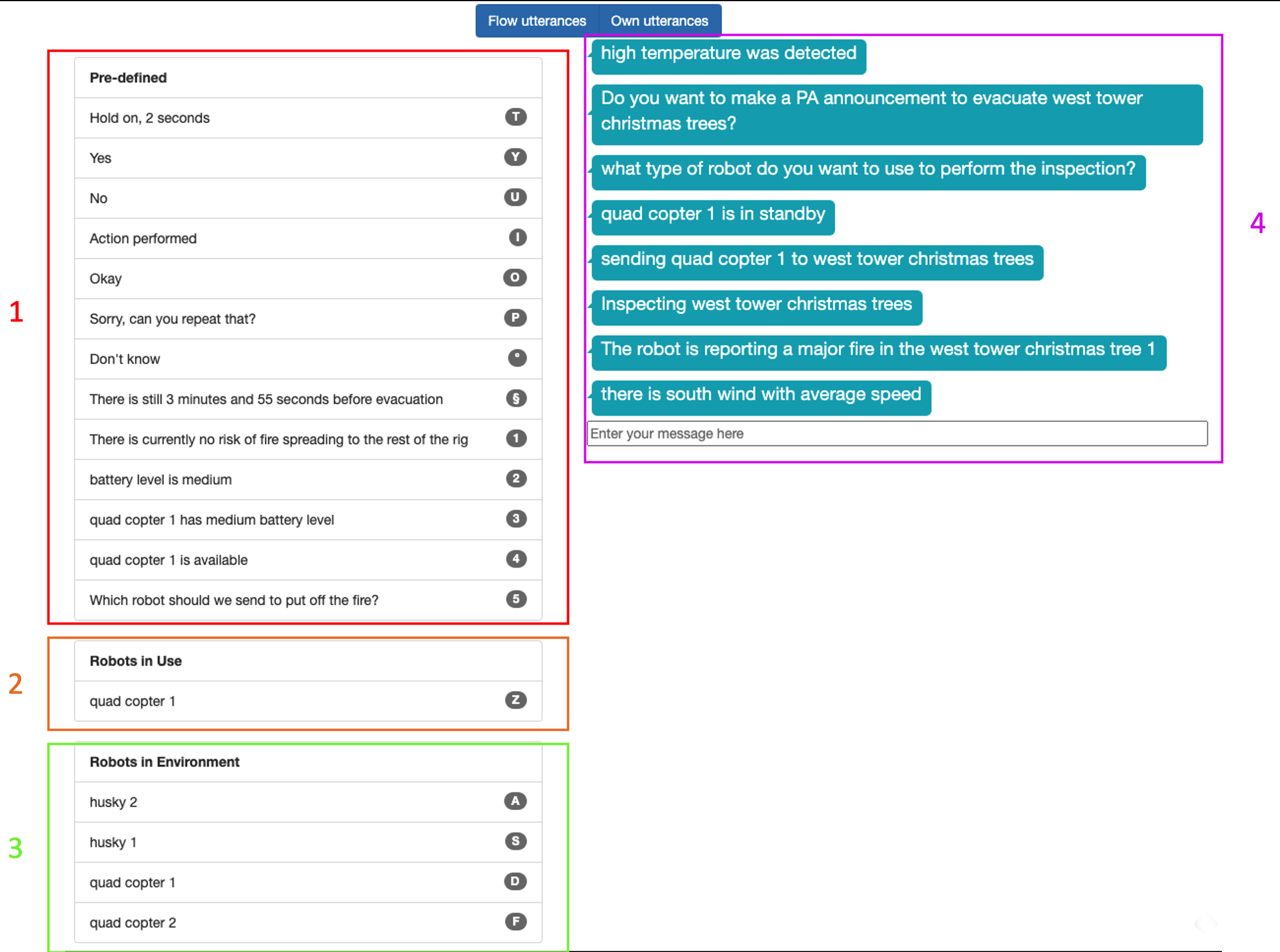}
        \caption{Lab wizard interface: (1) Prompts available, (2) robots in use, (3) robots available, (4) chat window. Details on the wizard interface can be found in \citet{Lopes2019HRI}.}
        \label{fig:wizard_lab}
    \end{subfigure}
    ~
    \begin{subfigure}[b]{0.45\textwidth}
        \includegraphics[width=\textwidth]{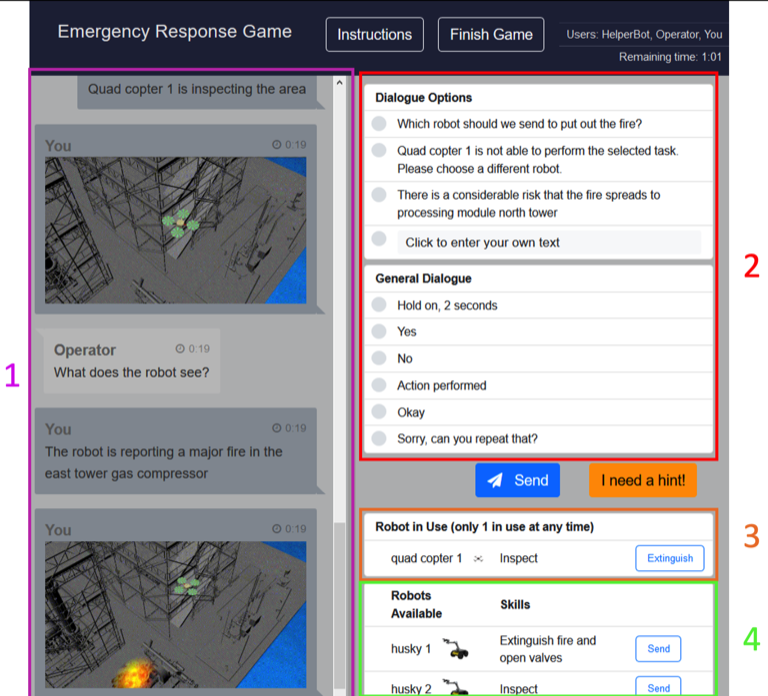}
        \caption{MTurk wizard interface: (1) Chat window, (2) Prompts available including free text box, (3) robots in use, (4) robots available.}
        \label{fig:wizard_mturk}
    \end{subfigure}
    \\
    \begin{subfigure}[b]{0.43\textwidth}
        \centering
        \includegraphics[width=\textwidth]{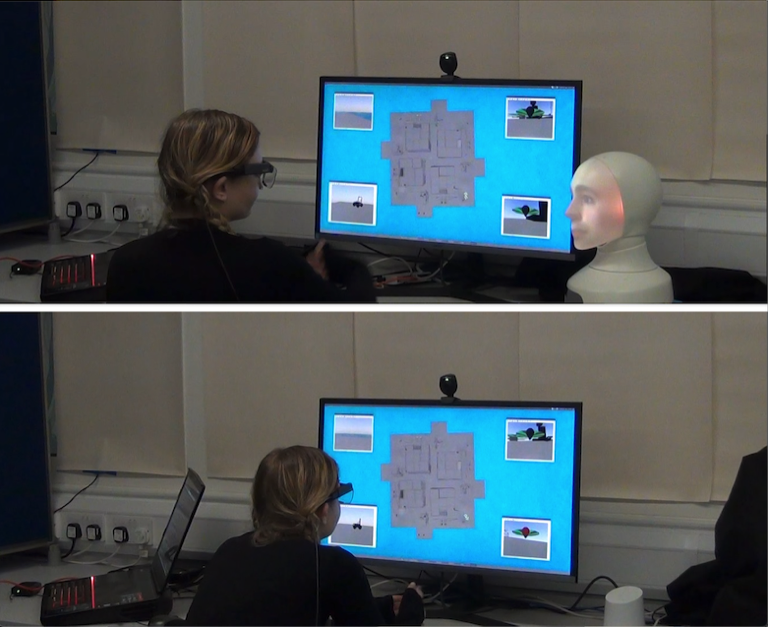}
        \caption{Lab operator interface with simulator: with Furhat (above) and Google Home (below).}
        \label{fig:user_lab}
    \end{subfigure}
    ~~~~~~~~
    \begin{subfigure}[b]{0.45\textwidth}
        \centering
        \includegraphics[width=\textwidth]{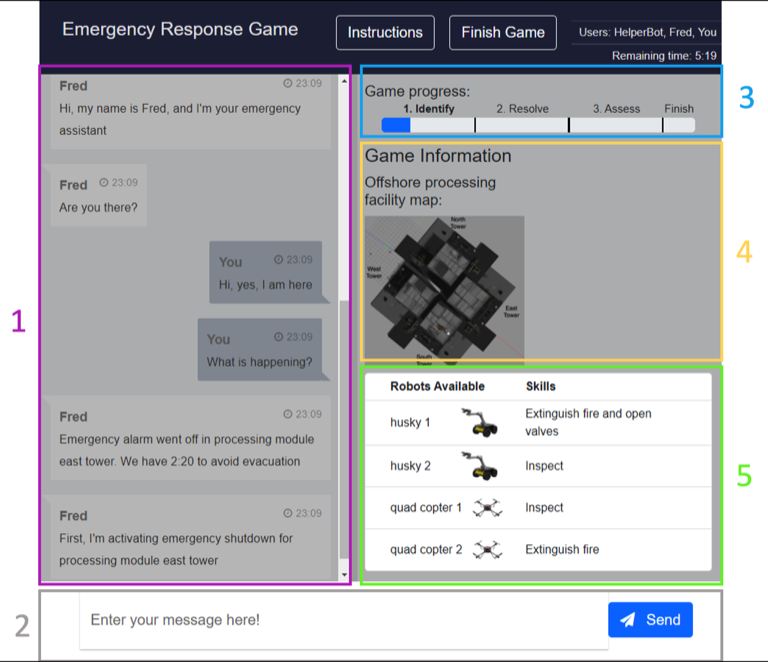}
        \caption{MTurk operator interface: (1) Chat window, (2) input text box, (3) mission progress bar, (4) top-view of facility, (5) robots available.}
        \label{fig:user_mturk}
    \end{subfigure}
    \caption{Interfaces used in each of the data collections.}
    \label{fig:interfaces}
\end{figure}

The participant who had the operator role was always able to speak or chat freely, whereas the wizard had to choose from a set of template-based utterances generated from a finite state automaton inspired by real practices from the energy sector. In addition, the wizard had a few pre-defined utterances such as \textit{Yes}, \textit{No} or utterances which could be used to manage the flow of the interaction such as \textit{Hold on}. As a last resort, wizards could type in an utterance of their own when there was no good fit amongst the options available. We collected data from two sources: a controlled lab study and a crowd-sourced task on Amazon Mechanical Turk. The interfaces used both for wizard and user are shown in Figure \ref{fig:interfaces}.

The main differences between the different collection methods were the following:

\begin{itemize}
    \item \textbf{Lab data}: one of the authors acted as the wizard in the task and participants were recruited from the university campus. Their task was to collaborate with a wizarded system, i.e. a wizard playing the role of an emergency assistant, embodied either by a social robot (a Furhat robot\footnote{\url{www.furhatrobotics.com}. See Figure \ref{fig:user_lab}.}) or a smart speaker to resolve the emergency. In both cases, the task was the same. The only difference was the emergency, which was placed in a different area of the off-shore energy platform in each condition. To help them through the process, subjects had access to a visualisation of the scenario, where the interaction was taking place (see Figure \ref{fig:user_lab}). They would interact through speech. The wizard could hear the subjects speaking and had access to a video stream of the participant and the scenario (see Figure \ref{fig:wizard_lab} for the wizard interface). We found no significant differences in the two conditions in terms of number of turns in the dialogue and number of words in participant utterances. Thus, dialogues collected in both conditions were treated as a single dataset.

    \item \textbf{Crowd-sourced data}: we paired crowd-workers together in a chat room where one of them would act as the wizard (playing the emergency assistant) in the interaction. We used a platform suited for this type of data collection so that the wizard had several predefined dialogue utterances available to use in a text-based chat. These would update as the task progressed and could trigger robots or events in the simulation, which would appear in the chat as GIFs (Figure \ref{fig:wizard_mturk} and \ref{fig:user_mturk}). Refer to \citet{chiyah-garcia-etal-2020-crwiz} for more information about the platform and this data collection\footnote{Instructions given to the workers are available at  \href{https://github.com/zedavid/TheLabVsTheCrowd}{\texttt{https://github.com/zedavid/TheLabVsTheCrowd}}. Workers were paid between \$1.4 and \$1.6 for a 10-minute HIT (equal to \$8.4/hr and \$9.6/hr), which was above the Federal minimum wage in the US (\$7.25/hr) at the time of the experiment.}.
\end{itemize}

From the lab collected data (Lab dataset), 3 dialogues had to be removed from the dataset due to missing log files, which reduced the number of usable dialogues in this setting to 63. The dialogues collected through Amazon Mechanical Turk (MTurk dataset) were manually filtered to remove those where the workers were not partaking in the task, which left a total of 147 dialogues.

Due to the different collection methods, the datasets differ in their proportion of interaction statistics, dialogue acts and resolved emergencies. Table \ref{tab:interaction_features} shows statistics of both datasets. 
Participants had a fixed amount of time in which they had 
to inspect an area of the platform and then extinguish the fire by any means available. This required careful planning  
and coordination between the emergency assistant and the operator. 
Aside from an overall lower number of turns per dialogue for the MTurk, other interaction characteristics of both datasets remained relatively similar. An example of one of the dialogues from the MTurk dataset can be found in Table \ref{tab:example_dialog}.

\begin{table}[h]
  \caption{Interaction statistics of the dialogues in the Lab and crowd-sourced MTurk datasets. Each turn corresponds to an utterance, e.g. in Table \ref{tab:example_dialog}, Utterances U1 and S1 are two different turns. 
  }
  \label{tab:interaction_features}
  \centering
  \begin{tabular}{|l|c|c|}
    \hline
    \textbf{Dataset Statistics} & \textbf{Lab Mean (SD)} & \textbf{MTurk Mean (SD)}  \\
    \hline
    Number of Turns & 49.08 (13.83) & 24.53 (9.49) \\
    \hline
    Number of Operator Turns & 11.43 (7.95) & 6.92 (3.24) \\
    \hline
    Number of Emergency Assistant Turns & 37.65 (8.57) & 17.61 (7.98) \\
    \hline
    Operator Turn Length (Words) & 4.37 (3.38) & 4.56 (3.35) \\
    \hline
    Emergency Assistant \% Typed Utterances & 2.58\% (2.87\%) & 2.77\% (5.14\%) \\
    \hline
  \end{tabular}
\end{table}

\begin{table}[ht]
    \caption{Excerpt of the beginning of a dialogue collected from the MTurk dataset. 
    } 
    \centering
    \small
    \begin{tabular}{|p{0.55\textwidth}|}
        \hline
        
         \textbf{S1}: Hi, my name is Fred, and I'm your emergency assistant. Are you there? \\  
         \textbf{U1}: \textbf{Yes. Quadcopter 2 extinguish fire!  }\\
         \textbf{S2}: Emergency alarm went off in processing module east tower. We have 1 minute 41 seconds to avoid evacuation. First, I'm activating the emergency shutdown. \\
        \textbf{U2}: Quadcopter 1 go inspect. \\
        \textbf{S3}: Quadcopter 1 is flying to processing module east tower. The robot is inspecting the area. \\
        \hline
    \end{tabular}
    \label{tab:example_dialog}
\end{table}



\subsection{Dialogue Models}

To model the dialogue, we have chosen to use Hybrid Code Networks (HCN) \citep{williams-etal-2017-hybrid}. HCNs offer an elegant solution combining an LSTM \citep{Hochreiter1997LSTM} with domain-dependent rules and procedures. HCNs do not need very large amounts of data to achieve a reasonable performance and, as reported in the original work, between 100 and 200 dialogues might be sufficient to reach a performance plateau. The HCN described in the paper uses as input features a Bag-of-Words (BoW) representation of the user utterance, a one-hot vector representation of the previous action and a context-dependent feature vector. We followed the same approach but, in our case, to obtain the context vector, we extracted the context features from the frame semantics information provided by the model described in \citep{vanzo-etal-2019-hierarchical}, which was re-trained including data from our domain. 
 
We then combined the original HCN features with word embeddings and an action mask. The word embeddings (Embed) were extracted as in \citep{williams-etal-2017-hybrid}, i.e., taking the average of the word embeddings in an utterance. For this, we have used word2vec embeddings  \citep{mikolov2013distributed} trained with Google News. For the purpose of direct comparison with the original architecture, we will keep the `Action Mask' (AM) as the denomination for the actions that are permitted at the current time step, captured as a one-hot vector. Following the same method used to create the Action Mask in \citep{williams-etal-2017-hybrid}, utterances requiring the robot slot would only be offered when the operator had selected a robot to perform a given task (e.g. inform\_moving). 
In addition, given that there were 3 sub-tasks involved, we restricted actions available at each turn following protocols/procedures and basic causality rules. For instance, ``the emergency cannot be resolved if not identified first by inspection'', or ``the outcome for the inspection cannot be provided before the inspection is completed''. The dialogue in Table \ref{tab:example_dialog} shows an example where the crowd-sourced operator was breaking this protocol by attempting to extinguish the fire before the inspection was performed (Utterance U1). Here, we observe the system continuing the interaction following the 3-step procedure, waiting for the operator to order the inspection first, rather than extinguishing a fire that has not been confirmed. This kind of behaviour was not observed in the Lab setting. 

\section{Results}
\label{sec:results}

The results from evaluating the HCN models trained on the different datasets with various combination of features are presented in Table \ref{table:results}. We report experiments with the full MTurk dataset (MTurk Full) of 147 dialogues, and with a subset of 63 randomly-selected dialogues (MTurk Partial),  matching the number of Lab dialogues. The datasets were first separated into training and validation. Given the limited size of the datasets, we evaluated the model in a leave-subject-out procedure, i.e., for each fold we trained a model using data from $N-1$ subjects and tested on the subject, who was left out of the training set. The validation set remained frozen throughout the $N$-folds. 


For our emergency response domain, we frame the model's effectiveness through Turn Accuracy (TA).  
Turn Accuracy refers to whether the predicted dialogue act is correct, compared to the gold standard for each juncture in the dialogue. 
We did not require a balance between operator and emergency assistant turns, hence the model was able to output a sequence of several dialogue acts without an operator's message. Similar Turn Accuracy is used to evaluate models in the original HCN paper \citep{williams-etal-2017-hybrid}. 

In addition to Turn Accuracy, we report perplexity (PP). To compute perplexity, we trained a dialogue act n-gram model with a set of dialogues, which included 63 dialogues from each dataset. Using the n-gram model, we computed the perplexity of the sequence of dialogue acts in each generated dialogue. Perplexity should give an idea of how likely is the dialogue generated, when compared to the distribution of dialogue act sequences in the ground-truth dialogues. The results for Turn Accuracy and Perplexity are presented in Table \ref{table:results}. For data with no specified testing set, these are calculated using the leave-subject-out procedure described above and averaged across the $N$-folds.



\begin{figure}
    \centering
    \includegraphics[width=0.60\textwidth]{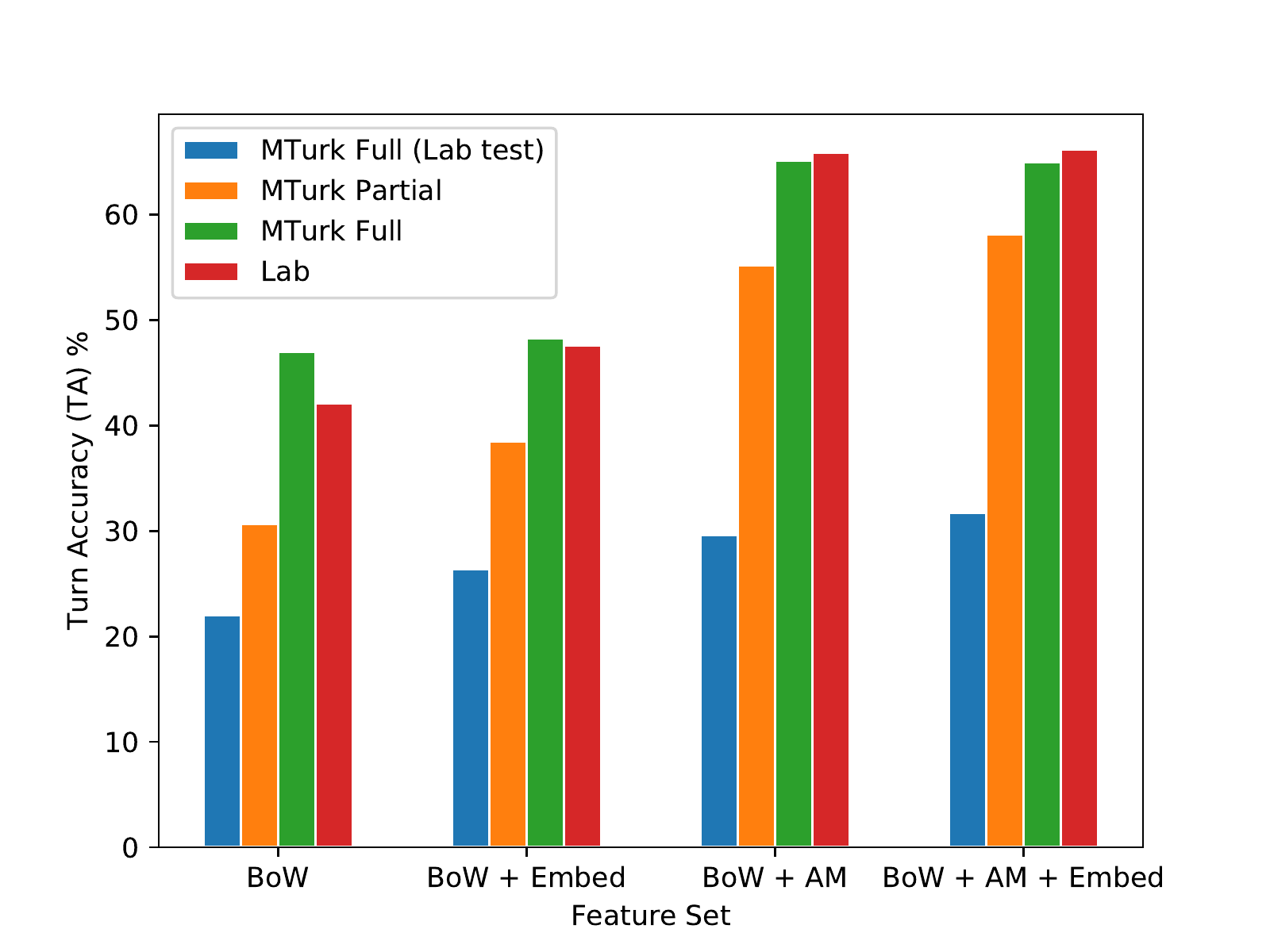}
    \caption{Turn Accuracy for the various system configurations and training and test sets. If the test set is not specified, it is the same category as the training set.}
    \label{fig:plot_turn_accruacy}
\end{figure}

In addition, we trained a model on the MTurk Full partition and tested in the Lab data. To minimise the impact of random sampling, we trained models 100 times with different splits of train and validation sets and report the average value for all the runs. Results from those runs can be found in the final row of each section in Table \ref{table:results}. In Figure \ref{fig:plot_turn_accruacy}, we see a plot with Turn Accuracy of the different training/testing procedures with various feature combinations.

As we can see from the results, when we compare models trained on the same sized datasets (Lab vs. MTurk Partial), there is a difference of performance (c.f. results for the full feature system: 58.1\% for MTurk Partial compared to  66.2\% for Lab). However, when we increase the MTurk dataset size to all the dialogues available (147), it reaches the same performance as the smaller Lab dataset (c.f. results for the full feature system: 65.0\% for MTurk Full compared to 66.2\% for Lab). Similar trends are seen with both model variants that use AM. With the two that do not use AM, the MTurk Full does slightly outperform the Lab trained system. However, the trend can still be seen across all models that the MTurk Full outperforms the MTurk Partial, matching or slightly beating the Lab model performance.
Note that the action mask was the most useful feature with respect to increasing the Turn Accuracy. 
This is because the Action Mask restricts the actions to those that adhere to the rules and procedures of the tasks, as is the nature of the hybrid technique used here. 


As a final set of results, we investigated how well pre-trained models would perform on our very specialised task. Thus, we followed the implementation proposed in \citep{wu2020tod} for response generation and fine-tuned our models using the same training and development partitions used during the training of the MTurk Full models and tested on the training Lab data. Results are shown in Table \ref{table:pretrained_models}. As we can see from the results, the best performance amongst pre-trained models (10.9\%) is well below the results reported in Table \ref{table:results} for models trained on the same Train/Test sets (22.0\% to 31.7\%). 


\begin{table}
\vspace{-0.3cm} 
\caption{Results from using a varying number of features for the HCN model with the different training/test datasets. The table shows Turn Accuracy (TA) and Perplexity (PP).  Dialogue features include:  Action Mask (AM), Bag of Words (BoW) and the use of embeddings in the model (Embed).}
  \label{table:results}
  \centering
  \begin{tabular}{|l|l|l|c|c|}
    \hline
    \textbf{Features} & \textbf{Training} (\# dialogues) & \textbf{Testing} (\# dialogues) & \textbf{TA} & \textbf{PP} \\
    \hline
    
    \multirow{3}{*}{BoW} &  MTurk Partial (59) & - & 30.7\% & 1.1264  \\
                         &  MTurk Full (138) & - & 47.0\% & 1.1479 \\
                         &  Lab  (59) & - & 42.1\%  & 1.0997  \\
                         & MTurk Full (138) & Lab (59) & 22.0\% & 1.0583 \\
    \hline
    \multirow{3}{*}{BoW + Embed} & MTurk Partial (59) & - & 38.5\% & 1.1236   \\
    & MTurk Full (138) & - & 48.3\% & 1.1369  \\
    & Lab (59) & - & 47.6\% & 1.0762  \\
    & MTurk Full (138) & Lab (59) & 26.4\% & 1.0441 \\
    \hline
    \multirow{3}{*}{BoW + AM} &  MTurk Partial (59) & - & 55.2\% & 1.1528  \\
                              & MTurk Full (138) & - & 65.1\% & 1.1695 \\
                              & Lab (59) & - & 65.9\% & 1.1172  \\
                              & MTurk Full (138) & Lab (59) & 29.6\% & 1.0325 \\
    \hline
    \multirow{3}{*}{BoW + AM + Embed} & MTurk Partial (59) & - & 58.1\% & 1.1412 \\
                                      & MTurk Full (138) & - & 65.0\% & 1.1676 \\
                                      & Lab (59) & - & \textbf{66.2\%} & 1.1227 \\
                                      & MTurk Full (138) & Lab (59) & 31.7\% & 1.0605 \\

    \hline
  \end{tabular}
\end{table}

\vspace{-0.3cm}

\begin{table}
  \caption{Results from using popular pre-trained models fine-tuned with all the MTurk dialogues and evaluated on Lab dialogues.}
  \label{table:pretrained_models}
  \centering
  \begin{tabular}{|l|l|l|c|}
    \hline
    \textbf{Model} & \textbf{Fine-Tuning} (\# dialogues) & \textbf{Testing} (\# dialogues) & \textbf{TA} \\
    \hline
    BERT \citep{Devlin2019}            & MTurk Full (147) & Lab (59) & 7.66\% \\
    ToDBERT \citep{wu2020tod}          & MTurk Full (147) & Lab (59) & 8.15\% \\
    DialoGPT \citep{zhang2019dialogpt} & MTurk Full (147) & Lab (59) & 10.9\% \\
    GPT2 \citep{radford2019language}   & MTurk Full (147) & Lab (59) & 8.19\% \\
    \hline
  \end{tabular}
\end{table}

\section{Discussion}
\label{sec:dicussion}




Results in Table \ref{table:results} show that the model trained only with Lab data outperformed the model trained with the complete dataset on (MTurk Full), with all the features included. When comparing the performance of models trained with the same number of dialogues (Lab vs MTurk Partial), the model trained with Lab data clearly outperformed the model trained with the MTurk data. This is not surprising, since the Lab data was collected with a single wizard, who had mastered the task. The wizard behaviour is more consistent than the behaviour of crowd-workers, who had little time to familiarise themselves with the task. The perplexity scores also confirm that dialogues trained with Lab data unfold in a more predictable way when compared with crowd-sourced data.

A fine-grain analysis revealed that the outputs of models trained with MTurk data tend to rush the dialogue, minimising the number of robot status updates (e.g. battery level) 
and using fewer non-task based dialogue acts, such as ``Hold on 2 seconds''. This pattern perhaps reflects the crowd-sourced worker's tendency to streamline tasks. However, even given the time constraint of the task, the above-mentioned dialogue acts (robot status and holding dialogue acts)
are important contributions to the dialogue, especially in terms of managing the user's confidence and stress levels. This effect could be due to the fact that face-to-face interactions require some turn-taking management, which is not reproducible in text-based dialogues.

Furthermore, the set-up used in the lab data collection is as close as one can get to the end application. As can be seen from Figure \ref{fig:interfaces}, the set-ups are significantly different. In the lab setting, participants are immersed in the scene in an operations-like environment, unlike in the crowd-sourced scenario where the interaction takes place through a chat window on the crowd-worker's computer. 
Clearly, the lab setting is more appropriate for tasks such as the emergency response task described here, where situation awareness and full user engagement are vital to replicate the conditions of the end application \citep{Robb2018}. Our results suggest that with a much smaller amount of data collected in very controlled conditions the model learns faster and more accurately. While this level of control might hinder the performance of the model when dealing with outliers, this seems not to be the case for our domain, where we expect participants to be highly knowledgeable of the task and compliant with the safety protocols that should be followed to complete the task successfully. 




\section{Conclusion and Future work}
\label{sec:conclusion}
Future work will involve investigating further the use of pre-trained models for specific-task based systems and the extent to which they can be used to bootstrap models for highly specific tasks such as our own. We acknowledge that the datasets in this study are small compared to datasets used to train state-of-the art neural models. This is one reason why we used the HCN method in this study as it has been shown to work well with small amounts of data \citep{williams-etal-2017-hybrid}. One future direction would be to duplicate such a study with a dataset of a similar size to MultiWOZ and explore further fine-grained increases in data size for training. This, however, will be very challenging and costly to collect lab data to match the size of the crowd-sourced data. With more in-lab data to retrain the models, we would plan to run a further in-depth systematic comparison of a variety of dialogue modelling approaches, for example using the methodology proposed in \citep{ultes-maier-2020-similarity}. Finally, to investigate the single-wizard impact, we are aiming to run a crowd-sourced data collection with a single wizard and repeat the experiments done in this paper. 

In this paper, we present a study comparing how different approaches to data collection may impact a hybrid neural dialogue model performance. Results suggest that, for our domain, models trained with small sets of lab-collected data outperform models trained with larger crowd-sourced datasets and pre-trained models. Given the nature of the domain, focusing on smaller lab data collections in realistic settings will likely be the best way to rapidly improve the model. However, the challenge to improve crowd-sourced data collection, making them as close as possible to the end application, still remains.

\begin{ack}

This work was supported by the EPSRC-funded ORCA Hub (EP/R026173/1, 2017-2021) and UKRI Trustworthy Autonomous Systems Node in Trust (EP/V026682/1, 2020-2024). Chiyah Garcia’s PhD is funded under the EPSRC iCase EP/T517471/1 with Siemens. We thank the anonymous reviewers for their insightful comments that helped improve this paper. We would like to thank Dr. David Robb, Dr. Muneeb Ahmad and Prof. Katrin Lohan for their work and input in the lab data collection.


\end{ack}

\bibliographystyle{plainnat}
\bibliography{neurips_2020}

\end{document}